\documentclass{article}
\usepackage[utf8]{inputenc}
\usepackage[english]{babel}
\usepackage{graphicx}
\usepackage{subfig}
\usepackage{hyperref}
\usepackage{fancyhdr}
\usepackage[sort, numbers]{natbib}

\begin{document}

\begin{center}
  \LARGE Adaptation of Deep Bidirectional Multilingual Transformers for Russian Language
\end{center}
\vskip 2em
\begin{center}
  \large \lineskip .75em%
  \begin{tabular}[t]{c}
    Yuri Kuratov \textsuperscript{1} \and Mikhail Arkhipov \textsuperscript{1} \\ \\
 \and
  \textsuperscript{1}  Neural Networks and Deep Learning Lab, \\
  Moscow Institute of Physics and Technology \\
  \and
  {\tt\small yurii.kuratov@phystech.edu~~~arkhipov.mu@mipt.ru}
  \end{tabular}\par
\end{center}

\renewcommand{\abstractname}{Abstract}
\begin{abstract}
The paper introduces methods of adaptation of multilingual masked language models for a specific language. Pre-trained bidirectional language models show state-of-the-art performance on a wide range of tasks including reading comprehension, natural language inference, and sentiment analysis. At the moment there are two alternative approaches to train such models: monolingual and multilingual. While language specific models show superior performance, multilingual models allow to perform a transfer from one language to another and solve tasks for different languages simultaneously.
This work shows that transfer learning from a multilingual model to monolingual model results in significant growth of performance on such tasks as reading comprehension, paraphrase detection, and sentiment analysis. Furthermore, multilingual initialization of monolingual model substantially reduces training time. Pre-trained models for the Russian language are open sourced.

\end{abstract} 

\selectlanguage{english}
\section{Introduction}

A large amount of work is devoted to unsupervised pre-training of neural networks on a variety of Natural Language Processing (NLP) tasks. Unsupervised pre-training shows significant improvements in almost every NLP task \cite{mikolov2013w2v, dai2015semi, peters2018elmo, devlin2018bert}. 

At the moment one of the best performing models for unsupervised pre-training is BERT \cite{devlin2018bert}. This model is based on Transformer \cite{vaswani2017attention} architecture and trained on a large number of unlabeled texts from Wikipedia to solve Masked Language Modelling task. It shows state-of-the-art results on a wide range of NLP tasks in English. 

Currently, there are publicly available two monolingual English and Chinese models and a single multilingual model. It was previously reported that monolingual models performance is significantly better than multilingual ones\footnote{\url{https://github.com/google-research/bert/blob/master/multilingual.md\#results}}. 

In the present work, we consider the possibility of multilingual to monolingual transfer. We use Russian as a target language for transfer. We show that it is possible to train the monolingual model using multilingual initialization.

To show this, we evaluated the multilingual model on a number of common NLP tasks from the target language. The model trained in a monolingual setting achieves substantially better performance compared to the multilingual model.

\section{Model Architecture}

In the present work, we use BERT \cite{devlin2018bert} model for all our experiments. The model is a Transformer \cite{vaswani2017attention} encoder. The basic building blocks of the model is Self-Attention. The model was trained on the Masked Language Modelling and next sentence prediction tasks. We refer readers to check BERT original paper for details about the model\cite{devlin2018bert}. 

We used 12 layers (Transformer blocks) version of BERT with self-attention hidden size 768, feed-forward hidden size 3072, and 12 self-attention heads. This setting corresponds to BERT\textsubscript{BASE} model from \cite{devlin2018bert}. Task-specific layers were trained according to the BERT paper.

\section{Language transfer}

In this work, we consider the transfer of multilingual BERT model to monolingual. Authors of \cite{devlin2018bert} showed that monolingual models show superior performance compared to multilingual one. 
% However, training of such model from scratch on the target language is highly computationally expensive. 
Furthermore, the BERT model uses the subword segmentation algorithm\cite{sennrich2015subwordnmt} to cope with large vocabulary problem. Multilingual models use only a small part of the entire vocabulary for a single language. It results in much longer sequences after tokenization compared to the monolingual model. Since the Transformer model has quadratic computational complexity in terms of input sequence length, it is highly undesirable.

We investigated the possibility of using the multilingual model as initialization for the monolingual model. The basic idea is to use knowledge about target language that already captured during multilingual training. It also is known that training model using data from multiple languages can significantly improve the performance of the model \cite{mulcaire2018polyglot}.
We used the multilingual model from BERT repository\footnote{https://github.com/google-research/bert}.
This model was trained on one hundred languages with largest Wikipedias.
The target language is Russian. All parameters of the model except word embeddings were initialized from the multilingual model \cite{mulcaire2018polyglot}. 

The new subword vocabulary was obtained using subword-nmt\footnote{https://github.com/rsennrich/subword-nmt}.
Training of the subword vocabulary was performed on the Russian part of Wikipedia and news data.
The part of Wikipedia data was around 80\%. The result of this step is a new monolingual Russian subword vocabulary. This vocabulary contains longer Russian words and subwords compared to multilingual one.  

New word embeddings matrix was obtained by assembling monolingual embeddings from multilingual. Namely, embeddings of all tokens from the intersection of multilingual and monolingual vocabulary were left without any changes. The same for special tokens like [UNK] or [CLS]. We replaced all tokens from outside the intersection with tokens from the monolingual vocabulary. These tokens are mostly longer subword units which are combinations of shorter units present in the intersection. New tokens are initialized with the mean value of embeddings from the intersection. For example, there are tokens 'bi' and '\#\#rd' in the intersection of vocabularies, where '\#\#' stands for the continuation of the word. There is also a token 'bird' present in the monolingual vocabulary and absent in the multilingual vocabulary. The embedding of 'bird' is initialized as the mean value of the embeddings 'bi' and '\#\#rd'.

The model with reassembled vocabulary and embeddings matrix was trained on the same data that was used for building of the monolingual vocabulary. The following hyperparameters were used for training:
\begin{itemize}
    \item batch size: 256
    \item learning rate: $2 \cdot 10^{-5}$
    \item optimizer: Adam
    \item L2 regularization: $10^{-2}$
\end{itemize}

% http://files.deeppavlov.ai/deeppavlov_data/bert/rubert_cased_L-12_H-768_A-12_v1.tar.gz
The monolingual Russian model\footnote{\url{http://files.deeppavlov.ai/deeppavlov_data/bert/rubert_cased_L-12_H-768_A-12_v1.tar.gz}} is available as a part of the \href{https://github.com/deepmipt/deeppavlov/}{DeepPavlov library}\footnote{https://github.com/deepmipt/deeppavlov/}.

\section{Tasks description}
We have chosen three tasks to evaluate our approach: paraphrase identification, sentiment analysis, and question answering. We briefly describe them in this section.

    \subsection{Paraphrase Identification with ParaPhraser}
    ParaPhraser \cite{pivovarova2017paraphraser} is a dataset for paraphrase detection in Russian language. Two sentences are paraphrases if they have the same meaning. This dataset consists of 7227/1924 train/test pairs of sentences which are labeled as precise paraphrases, near paraphrases or non-paraphrases. One approach for paraphrase identification is a binary classification: first class is precise and near paraphrases, second class - non-paraphrases.
    
    \subsection{Sentiment Analysis with RuSentiment}
    RuSentiment \cite{rogers2018rusentiment} is a dataset for sentiment analysis of posts from VKontakte (VK), the most popular social network in Russia. Realised in 2018, it became one of the largest sentiment datasets for Russian language with 30521 posts. Each post is labeled with one of the five classes. The informal language presented in RuSentiment dataset makes it more challenging for our model, trained on Wikipedia and news articles.
    
    \subsection{Question answering with SDSJ Task B}
    As part of Sberbank Data Science Journey\footnote{https://sdsj.sberbank.ai/} 2017 was held a competition with two tasks. Task B was inspired by Stanford Question Answering Dataset (SQuAD) \cite{rajpurkar2016squad}. Organizers collected about 50,000 (train and development set) questions and contexts, where the answer is always a span from corresponding context. SQuAD dataset encouraged community to develop sophisticated and effective neural architectures such as BiDAF \cite{seo2016bidirectional}, R-NET \cite{wang2017gated}, Mnemonic Reader \cite{hu2017reinforced}. All this models are based on attention mechanisms \cite{bahdanau2014neural,luong2015effective} and building joint context-question representation.

\section{Results}
We evaluated BERT multilingual model and BERT trained with our approach (RuBERT) on three tasks: paraphrase identification, sentiment analysis, and question answering. All reported results were obtained by averaging across 5 runs.

\begin{table}[!ht]
\centering
\begin{tabular}{l|l|l}
\hline
model                            & F-1   & Accuracy \\ \hline
Neural networks \cite{pivovarova2017paraphraser}  & 79.82 & 76.65    \\ \hline
Classifier + linguistic features \cite{pivovarova2017paraphraser} & 81.10 & 77.39    \\ \hline
Machine Translation + Semantic similarity \cite{kravchenko2017paraphrase} & 78.51 & 81.41    \\ \hline\hline
BERT multilingual                & $85.48\pm0.19$ & $81.66\pm0.38$    \\ \hline
RuBERT                           & $\textbf{87.73}\pm0.26$ & $\textbf{84.99}\pm0.35$    \\ \hline
\end{tabular}
\caption{ParaPhraser. We compare BERT based models with models in non-standard run setting, when all resources were allowed.}
\label{tab:paraphraser}
\end{table}

\begin{table}[ht!]
\centering
\begin{tabular}{l|l|l|l}
\hline
model               & F-1   & Precision & Recall \\ \hline
Logistic Regression \cite{rogers2018rusentiment} & 68.84 & 69.53     & 69.46  \\ \hline
Linear SVC \cite{rogers2018rusentiment}         & 68.56 & 69.46     & 69.25  \\ \hline
Gradient Boosting \cite{rogers2018rusentiment}   & 68.48 & 69.63     & 69.19  \\ \hline
NN classifier \cite{rogers2018rusentiment}       & 71.64 & 71.99     & 72.15  \\ \hline \hline
BERT multilingual   & $70.82\pm0.75$ & -         & -      \\ \hline
RuBERT              & $\textbf{72.63}\pm0.55$ & -         & -      \\ \hline
\end{tabular}
\caption{RuSentiment. We used only randomly selected posts (21,268) subset for training.}
\label{tab:rusentiment}
\end{table}

\begin{table}[ht!]
\centering
\begin{tabular}{l|l|l}
\hline
model                 & F-1 (dev) & EM (dev) \\ \hline
R-Net from DeepPavlov \cite{burtsev2018deeppavlov} & 80.04     & 60.62    \\ \hline \hline
BERT multilingual     & $83.39\pm0.08$     & $64.35\pm0.39$    \\ \hline
RuBERT                & $\textbf{84.60}\pm0.11$     & $\textbf{66.30}\pm0.24$    \\ \hline
\end{tabular}
\caption{Results on question answering with SDSJ Task B. Models performance was evaluated on development set (public leaderboard subset).}
\label{tab:squad}
\end{table}

SDSJ Task B and ParaPhraser datasets share the same domain with data, which we used for training RuBERT. The RuSentiment dataset is based on posts from a social network and shows to be more challenging for RuBERT. As result, we can see only 1 F-1 point improvement from previous state of the art for RuSentiment in Table \ref{tab:rusentiment}, comparing to 4-6 F-1 points improvement on SDSJ Task B and ParaPhraser (results in Table \ref{tab:paraphraser} and Table \ref{tab:squad}).

\subsection{Vocabulary comparison}
\begin{figure}[ht]%
    \centering
    \subfloat[BERT multilingual]{{\includegraphics[width=5cm]{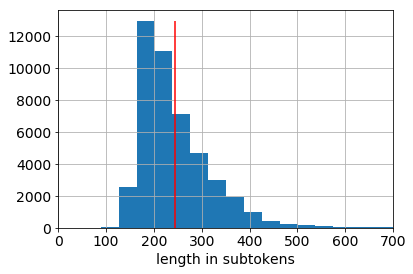} }}%
    \qquad
    \subfloat[RuBERT]{{\includegraphics[width=5cm]{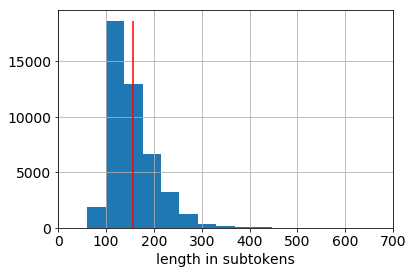}}}%
    \caption{Distribution of lengths in subtokens of contexts with their questions (SDSJ Task B dataset). Red vertical lines represent mean values.}%
    \label{fig:vocab_hist}%
\end{figure}

BERT multilingual and RuBERT have the same size of vocabulary (about 120k subtokens), but RuBERT vocabulary was built especially for Russian language. Figure \ref{fig:vocab_hist} shows that RuBERT model allows to reduce mean sequence length in $1.6$ times in subtokens, what makes possible to increase batch size or feed longer texts to the model, comparing to BERT multilingual.

\subsection{Training dynamics}
\begin{figure}[ht!]
\centering
\includegraphics[width=9cm]{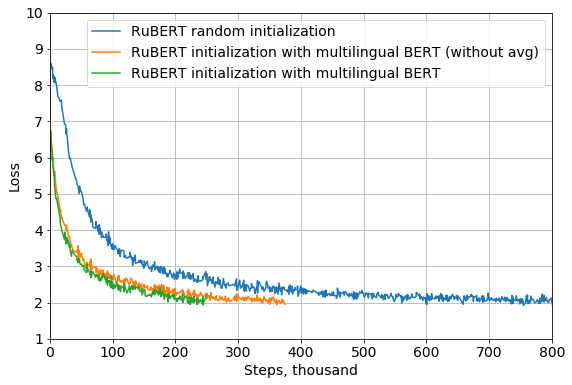}
\caption{Models training dynamics to get to the same value of loss.}
\label{fig:loss}
\end{figure}

In this section we compare training BERT model for Russian language from scratch (random initialization) and initialized with BERT multilingual. Figure \ref{fig:loss} shows that BERT multilingual initialization helps to converge faster: about 800 thousand steps is required for random initialized model to get the same loss as at 250 thousand step of multilingual initialization. It takes about two days to train for 250 thousand steps (on Tesla P100 x 8), so it helped us to save six days of computational time. Proposed averaging of new subtokens in vocabulary also has positive effect on the rate of convergence (instead of averaging we could take random initialization for new subtokens).

\section{Conclusion}

In this work, we have shown that Transformer network pre-trained on the multilingual Masked Language Modelling task significantly improves performance on a number of Russian NLP tasks compared to existing solutions. Furthermore, language-specific unsupervised training with multilingual initialization results in even better improvements. Pre-trained models for the Russian language are open sourced, as well as code to reproduce our results as part of DeepPavlov library.

\section{Acknowledgments}
This work was supported by National Technology Initiative and PAO Sberbank project ID 0000000007417F630002.

\end{document}